\documentclass[pdflatex]{sn-jnl}

\usepackage{balance} 
\usepackage{amsmath}
\usepackage{microtype}
\usepackage{graphicx}
\usepackage{scalerel}
\usepackage{subcaption}
\usepackage{stfloats}
\usepackage{enumitem}
\usepackage{tabu}
\usepackage{diagbox}
\usepackage{dirtytalk}
\usepackage[numbers,sort&compress]{natbib}

\jyear{2023}%

\DeclareMathOperator*{\argmax}{argmax}

\makeatletter
\newcommand*{\inlineequation}[2][]{%
  \begingroup
    \refstepcounter{equation}%
    \ifx\\#1\\%
    \else
      \label{#1}%
    \fi
    \relpenalty=10000 %
    \binoppenalty=10000 %
      #2%
    ~\@eqnnum
  \endgroup
}
\makeatother

\begin{document}

\title[QGNN: Value Function Factorisation with Graph Neural Networks]{QGNN: Value Function Factorisation with \\ Graph Neural Networks}

\author[1]{\fnm{Ryan} \sur{Kortvelesy}}\email{rk627@cam.ac.uk}

\author[1]{\fnm{Amanda} \sur{Prorok}}\email{asp45@cam.ac.uk}

\affil[1]{\orgdiv{Department of Computer Science and Technology}, \orgname{University of Cambridge}, \orgaddress{\city{Cambridge}, \country{UK}}}

\abstract{In multi-agent reinforcement learning, the use of a global objective is a powerful tool for incentivising cooperation. Unfortunately, it is not sample-efficient to train individual agents with a \textit{global} reward, because it does not necessarily correlate with an agent's individual actions. This problem can be solved by factorising the global value function into \textit{local} value functions. Early work in this domain performed factorisation by conditioning local value functions purely on local information. Recently, it has been shown that providing both local information and an encoding of the global state can promote cooperative behaviour. In this paper we propose QGNN, a value factorisation method which uses a graph neural network (GNN) based model. The multi-layer message passing architecture of QGNN provides more representational complexity than models in prior work, allowing it to produce a more effective factorisation. QGNN also introduces a permutation invariant mixer which is able to match the performance of other methods, even with significantly fewer parameters. We evaluate our method against several baselines, including QMIX-Att, GraphMIX, QMIX, VDN, and hybrid architectures. Our experiments include Starcraft, the standard benchmark for credit assignment; Estimate Game, a custom environment that explicitly models inter-agent dependencies; and Coalition Structure Generation, a foundational problem with real-world applications. The results show that QGNN outperforms state-of-the-art value factorisation baselines consistently.
}


\maketitle

\section{Introduction}

In multi-agent reinforcement learning, the task of utilising cooperation to achieve a global objective is of particular interest to the research community. A few applications include connected autonomous vehicles~\cite{hyldmar_Fleet_2019}, sensor networks \cite{SensorNets}, and modular control of actuators in a single robot \cite{ModControl}. In these scenarios, there is a clear global objective, but local rewards are often not defined. While it is possible to design local reward functions with reward shaping, doing so can lead to greedy behaviour \cite{MultiagentRewards}. 

One particularly promising solution is the use of centralised training with a decentralised (or communication-based) policy, where agents only have access to information communicated within a local neighbourhood \cite{POMDP}. This allows the global objective to be optimised without subjecting learning to the curse of dimensionality. However, the global reward cannot be directly used to train a policy, as the reward signal for any individual agent is confounded by the rest of the system \cite{MultiagentRewards}. In order to stabilise training, the global reward must be distributed into local rewards concomitant with agents' individual contributions. This is known as the credit assignment problem. 

While significant research has delved into the credit assignment problem, most methods rely on factorisation into local value functions conditioned exclusively on local information \cite{IQL, VDN, QMIX, QTRAN, GraphMIX, qplex, qrelation, raca}. However, in most problems, the global value function cannot be separated into \textit{independent} local value functions.

In this paper, we tackle value function factorisation with Graph Neural Networks (GNNs),
leveraging a learned multi-hop communication strategy to calculate local values. This method facilitates cooperative behaviours in environments where the optimal local value functions are interdependent.

\textbf{Contributions:}
\begin{itemize}[leftmargin=0.4cm,noitemsep,topsep=0pt]
    \item We present QGNN, a novel value factorisation method with a GNN-based agent model and a permutation invariant mixer.
    \item QGNN outperforms state-of-the-art baselines in three distinct experiments: Starcraft (the standard benchmark for credit assignment), Estimate Game (a custom environment which explicitly models inter-agent dependencies), and Coalition Structure Generation (a foundational problem with real-world applications).
    \item QGNN's graph structure allows users to model real-world constraints like communication range, encode information about known inter-agent dependencies, and incrementally transition from a non-communicative model to one that fully utilises communication. We experimentally demonstrate the importance of operating over a graph structure, and we perform an ablation analysis over graph connectivity by adjusting communication range.
    \item We provide the insight that representational complexity is more important inside a communication-based model than inside a mixer. This claim is grounded in theory and backed up by an ablation analysis over our architecture's components.
    \item We demonstrate that there are parallels between the manner in which GNNs compress information, the calculation of Shapley values, and value factorisation in model-mixer architectures. By introducing an experimental environment where the underlying local values are known, we introduce a new metric which allows different methods to be compared based on their ability to implicitly learn the true local values. Through this metric, we show that a GNN can \textit{implicitly} perform credit assignment.
    
\end{itemize}


\section{Background}
\label{sec:background}

Consider a cooperative multi-agent task, modelled by a DEC-POMDP \cite{POMDP} of the form $\langle \mathcal{S}, \mathcal{U}, P, R, \mathcal{O}, \mathcal{G}, n, \sigma, \gamma \rangle$. This denotes a system of $n$ agents with a global state $\mathbf{s} \equiv \{s_i\}_{i=1}^n \in \mathcal{S}^n$. 
Individual agents draw their observations $o_i \in \mathcal{O}$ from the observation function $o_i \sim \sigma(s_i)$. Each agent $i$ selects an action $u_{i} \in \mathcal{U}$, generating a joint action $\mathbf{u} \equiv [u_{i}]_{i=1}^{n} \in \mathcal{U}^n$. Over the course of an episode of length $T$, the action-observation history $\tau_{i} \in \mathcal{T} : (\mathcal{O} \times \mathcal{U})^T \times \mathcal{O}$ is stored. This local trajectory $\tau_{i}$ is shared with other agents through communication, which is governed by the underlying communication graph $\mathcal{G} \equiv \langle \mathcal{V}, \mathcal{E} \rangle$. Nodes $v_i \in \mathcal{V}$ represent agents, and edges $e_{ij} \in \mathcal{E}$ represent communication links. Each agent $i$ receives data from its neighbourhood $\mathcal{N}_i \equiv \{ v_i \} \cup \{v_j \, \vert \, e_{ij} \in \mathcal{E}\}$ to generate a joint trajectory $\boldsymbol{\tau}_{\mathcal{N}_i} \equiv [\tau_{i}]_{i \in \mathcal{N}_i}  \in \mathcal{T}^{\lvert \mathcal{N}_i \rvert}$. A policy uses this information  $\pi(\boldsymbol{\tau}_{\mathcal{N}_i}) : \mathcal{T}^{\lvert \mathcal{N}_i \rvert} \mapsto \mathcal{U}$ to produce an action $u_i$ for each agent. At each timestep $t$, the system receives a global reward $r_t = R(\mathbf{s_t},\mathbf{u_t}) : (\mathcal{S} \times \mathcal{U})^n \mapsto \mathbb{R}$. Then, the state transition function $P\left(\mathbf{s}_{t+1}\; \vert\; \mathbf{s}_t,\mathbf{u}_t\right) : \mathcal{S}^n \times \mathcal{S}^n \times \mathcal{U}^n \mapsto [0,1]$ dictates the probability of transitioning to a new joint state. The goal of the task is to find the policy $\pi$ which maximises the expected discounted return $\mathbb{E}_{\, \mathbf{s}_0 \sim \mathcal{S}, \mathbf{s}_{t+1} \sim P(\mathbf{s}_t, \mathbf{u}_t)}\left[\sum_{i=0}^{\infty} \gamma^i R(\mathbf{s}_{t+i}, \pi(\mathbf{s}_{t+i})) \right]$.

\sloppy Using reinforcement learning, the value of each joint state can be modelled with the Bellman operator $Q(\boldsymbol{\tau}_t,\mathbf{u}_t) = \mathbb{E}\left[ r_t + \gamma \max_{\mathbf{a}}Q(\boldsymbol{\tau}_{t+1},\mathbf{a}) \right]$. The task of value factorisation entails the decomposition of this global value $Q$ into local values $q_i$ for each individual agent $i$. In order to achieve globally optimal behaviour, it is necessary for the globally optimal action to maximise the local value functions. Appropriately, this constraint is called \textit{Individual-Global Max} (IGM) \cite{QTRAN}. In our work, we apply a relaxation to the original formulation, allowing the local q-values to be conditioned on the joint trajectory:

\begin{equation}
\argmax_{\mathbf{u}} Q(\boldsymbol{\tau}, \mathbf{u}) = 
\begin{pmatrix}
	\argmax_{u_1} q_1({\boldsymbol{\tau}_{\mathcal{N}_1}}, u_1) \\
	\vdots \\
	\argmax_{u_n} q_n({\boldsymbol{\tau}_{\mathcal{N}_n}}, u_n) 
\end{pmatrix}
\label{eq:IGM}
\end{equation}


\section{Related Work}
\label{sec:related_work}

In the early stages of research in multi-agent reinforcement learning, the notion of separating a global value function into local value functions emerged with an approach called \textit{Independent Q-Learning} (IQL) \cite{IQL}. While a naïve method would simply use the global reward to train a global value function, IQL maintains a separate value function for each agent. These local value functions are trained with the global reward, but they are conditioned upon local information. Since they can only represent the portion of the global reward which can be predicted with local information, the individual contributions of each agent can be isolated. However, this method is not sample-efficient because of the non-stationarity of the global reward signal---the portion of the signal which cannot be represented by local information appears to be noise.

In an attempt to address the non-stationarity problem, \textit{Value Decomposition Networks} (VDN) introduced the concept of simultaneously training all local value functions in an end-to-end manner with the use of a \say{mixer} \cite{VDN}. In VDN, the mixer is simply a summation, which aggregates the local q-value estimates to produce a global Q-value estimate. This formulation reduces instability because it splits the gradient of the global value loss among all of the local value networks. However, VDN only works well on the subset of problems in which the global value can be represented as a \textit{sum} of local values conditioned on local information. In subsequent work, QMIX provides the insight that it is not necessary to combine local q-values with a summation in order to satisfy the IGM condition \cite{QMIX}---it is sufficient to maintain a monotonicity constraint $\forall i, \frac{\partial Q}{\partial q_i} \geq 0$. With this relaxation, it is possible to use a more complex mixer which can capture a larger set of problems than VDN. 

Building on the representational complexity improvements of QMIX, many recent methods employ different mixing strategies. QTRAN \cite{QTRAN} seeks to represent non-monotonic value functions by employing optimisation objectives to satisfy the IGM condition~\eqref{eq:IGM}. QPLEX \cite{qplex} and QSCAN \cite{qscan}, on the other hand, apply a transformation to the individual q-values inside of the mixing network to condition them on global information.

There are several models which use GNNs for value factorisation: GraphMIX \cite{GraphMIX}, QRelation \cite{qrelation} and RACA \cite{raca}. These methods maintain locally conditioned models, but they use monotonic GNNs in the mixer to model relationships between agents. The key drawback of these methods is that they cannot represent inter-agent dependencies, because the local value functions are conditioned purely on local information. Given that the learned policies must be agnostic to the states of other agents, only limited coordination is possible.

One method that can model interactions on a macroscopic scale is COMA \cite{COMA}. In COMA, a global value function is used to calculate agent-wise advantages by subtracting a counterfactual baseline. However, due to the centralised nature of the global critic, COMA exhibits poor scalability and generalisation to new states. Furthermore, since COMA uses an actor conditioned on local information (like the aforementioned methods), the resulting policy cannot observe the states of other agents to produce cooperative behaviour.

Some methods use communication to improve performance in multi-agent reinforcement learning \cite{MADDPG, DGN, G2ANet, LA-QTransformer, TarMAC, CommNet, AI-QMIX}. The most popular of these methods is MADDPG \cite{MADDPG}, which uses communication at train time, but not at execution time. There are also similar methods which utilise communication at training time \textit{and} execution time, such as DGN \cite{DGN} and GA-Comm \cite{G2ANet}. However, all of these methods require local rewards, so they are not useful for the fully cooperative tasks tackled in this paper, which only provide a global reward.

There exist some communication-based approaches which tackle fully cooperative tasks, such as PIT/LA-QTransformer \cite{LA-QTransformer}, TarMAC \cite{TarMAC}, and CommNet \cite{CommNet}. Like MADDPG, PIT/LA-QTransformer utilises communication exclusively at training time, but it uses a model-mixer architecture to tackle problems with global rewards \cite{LA-QTransformer}. In contrast, TarMAC and CommNet use communication at training and execution time, but they do not perform value function factorisation. They directly train local policies with a global value function (TarMAC) or global reward (CommNet). Since they do not address credit assignment, the agents' actions are reinforced according to the entire team's performance, regardless of individual contributions. This leads to poor sample efficiency and an inability to scale to a large number of agents in complex tasks.

Coordination Graphs constitute another related field within MARL which utilises communication. Deep Coordination Graphs (DCG) learns utility $f(a_i \vert \tau_i)$ and payoff $f(a_i,a_j \vert \tau_i,\tau_j)$ with a training procedure similar to VDN (with the addition of pairwise value functions) \cite{DCG}. Then, through message passing, agents iteratively update their actions to maximise the global value. In Deep Implicit Coordination Graphs (DiCG), a GNN is used to generate local embeddings for each agent (which can be cast into actions and q-values). The shared local policy is directly trained with the global reward using an actor-critic method. The primary difference from our work is that DiCG only uses a GNN as a model (and does not use a mixer, like CommNet), whereas in our method the GNN acts as \textit{both} the model \textit{and} the mixer, implicitly performing credit assignment.

The most similar approach to ours is QMIX-Att, introduced in REFIL \cite{AI-QMIX}, as it is the only other value factorisation method which uses communication. While REFIL also introduces a new training procedure which involves the consideration of random neighbourhoods, the most relevant component of the method is their model, QMIX-Att, which applies multi-head attention to QMIX. There are several key differences between QGNN and QMIX-Att. First, while QMIX-Att applies attention globally, QGNN learns interdependencies on a graph structure with a multi-layer GNN, allowing it to perform hierarchical reasoning. Second, while QMIX-Att uses the QMIX mixer, QGNN implements a permutation-invariant mixer with fewer parameters, affording it better generalisation capabilities. To demonstrate these difference empirically, we use QMIX-Att as a baseline in our experiments.


\section{Method}

QGNN is composed of a model and a mixer. In prior work, most research focuses on the development of \textit{mixers} with increasing representational complexity \cite{VDN, QMIX, QTRAN, GraphMIX, qplex, qrelation, raca}. In these architectures, the function of the mixer is to distribute gradients to the local value functions---an ideal mixer would provide gradients which mimic training with independent local rewards. However, even with a perfect mixer, it is impossible to learn the correct local value function without communication. With value functions conditioned on purely local information, the \say{ground truth} local rewards appear to be non-stationary. This also makes it impossible to represent interdependent or non-monotonic value functions, where the optimality of a state or action is affected by non-local information (Appendix \ref{appendix:nonmonotonic}). In such scenarios, there exist no locally-conditioned value functions $q_i$ which can satisfy the IGM condition \eqref{eq:IGM}. However, if the local value functions are conditioned on non-local information, then there exist value functions $q_i$ such that the monotonicity constraint $\frac{\partial Q}{\partial q_i} \geq 0 \; \forall i \in \{1..N\}$ holds (Appendix \ref{appendix:nonmonotonic}).

The key insight of our work is that the model should take over the role of representing interdependent value functions. Hence, we design a model that can represent interdependence through communication within a multi-layer message-passing GNN architecture. Compared to QMIX-Att \cite{AI-QMIX} and other factorisation-based methods \cite{VDN, QMIX, QTRAN, GraphMIX, qplex, qrelation, raca}, QGNN shifts the complexity in the architecture from the mixer to the model. We provide further theoretical justification for this design choice in Appendix \ref{appendix:shapley}. The combined architecture of the QGNN model and mixer forms a single graph network, as defined in \cite{GraphNets}. The model represents the edge and node update steps, producing factorised local values. The mixer, in turn, represents the global update step, producing an estimate of the global value function. The success of our architecture makes a powerful statement, as it shows that GNNs for graph-level predictions can implicitly learn factorised local values simply by maintaining a monotonic global update step. Furthermore, it suggests that highly specialised existing models for value function factorisation can be replaced by existing GNN architectures.

\subsection{Model}

\begin{figure*}[t]
    \begin{subfigure}[b]{0.49\textwidth}
    	\centering
        \includegraphics[width=0.7\linewidth]{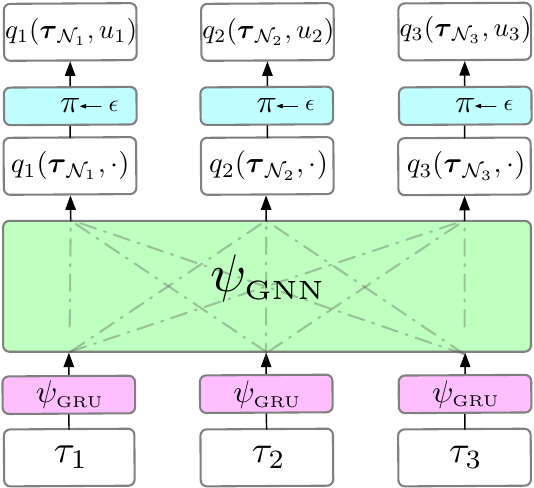}
    	\caption{QGNN model}
    	\label{fig:QGNN_model}
    \end{subfigure}
    \begin{subfigure}[b]{0.49\textwidth}
    	\centering
        \includegraphics[width=0.7\linewidth]{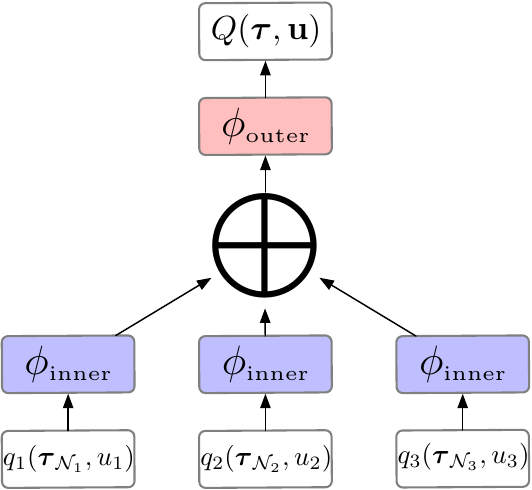}
    	\caption{QGNN mixer}
    	\label{fig:QGNN_mixer}
    \end{subfigure}
    \caption{The QGNN architecture for a fully connected graph of $n=3$ agents. The model encodes trajectories $\tau_i$ with a GRU $\psi_\text{\tiny{GRU}}$, and then uses a GNN $\psi_\text{\tiny{GNN}}$ to produce a local q-value for each action. Then, an $\epsilon$-greedy policy $\pi$ selects the action which maximises the local value. Finally, the local values $q_i$ are combined into a global value estimate $Q$ with the mixer. This involves the application of an ``inner'' network $\phi_\text{\tiny{inner}}$, generalised aggregation, and an ``outer'' network $\phi_\text{\tiny{outer}}$.}
    \label{fig:QGNN_architecture}
\end{figure*}

The QGNN model builds on top of the GRU-based models from prior work \cite{VDN, QMIX, GraphMIX}. The GRU-encoded trajectories $x_i = \psi_\text{\tiny{GRU}}(\tau_i)$ are passed into a GNN $\psi_\text{\tiny{GNN}}$, which produces q-value for each possible action for each agent $\{q_i({\boldsymbol{\tau}_{\mathcal{N}_i}}, a) \, \vert \, a \in \mathcal{U} \}_{i = 1}^n = \psi_\text{\tiny{GNN}}(\mathbf{x})$ (Fig. \ref{fig:QGNN_model}).

In practice, we implement $\psi_\text{\tiny{GNN}}$ with a 2-layer \textit{EdgeConv} GNN \cite{EdgeConv}. From a decentralised point of view, the values $q_i({\boldsymbol{\tau}_{\mathcal{N}_i}}, \cdot)$ for each possible action of agent $i$ are calculated: $\{q_i({\boldsymbol{\tau}_{\mathcal{N}_i}}, a) \, \vert \, a \in \mathcal{U} \} = f_\text{\tiny{critic}}(x_i^{(0)} \vert\vert \, x_i^{(2)})$, where $x_i^{(l+1)} = \sum_{j \in \mathcal{N}_i} f_\text{\tiny{edge}} \left( x_i^{(l)} \vert\vert \, x_j^{(l)}-x_i^{(l)} \right)$. In this formulation, $f_\text{\tiny{edge}}$ and $f_\text{\tiny{critic}}$ are multi-layer perceptrons and $x_i^{(0)} = x_i$. Finally, the joint action is selected by taking the argmax over each agent's q-value $\mathbf{u} = \{\mathrm{argmax}_{a \in \mathcal{U}} \, q_i({\boldsymbol{\tau}_{\mathcal{N}_i}}, a) \}_{i=1}^n$.

The number of GNN layers in the model can be increased to provide a larger receptive field and more representational complexity at the cost of additional computation. The connectivity of the graph, however, is determined by the problem---it can encode priors about known interactions, or physical constraints like communication range. In scenarios where the graph is unknown, the assumption of full connectivity can still provide performance benefits over non-communicating models (Sec. \ref{section:starcraft}).

\subsection{Mixer}

The input to the mixer consists of the q-values corresponding to the selected actions $\mathbf{q} = \{q_i({\boldsymbol{\tau}_{\mathcal{N}_i}}, u_i)\}_{i=1}^n$. These values $\mathbf{q}$ are used to produce a global value estimate. 

The QGNN mixer is analagous to the global update step in a GNN for graph regression, so we implement it with a custom global pooling operation \cite{GraphNets}. The mixer seeks to provide as much representational complexity as possible with a relatively simple architecture. To achieve this, we use Deep Sets \cite{DeepSets}, which serves as a universal approximator of set functions, which are symmetric, and therefore permutation-invariant---a property which follows from the Kolmogorov-Arnold representation theorem \cite{Kolmogorov}:
	
\begin{equation}
    Q(\boldsymbol{\tau},\mathbf{u}) = \phi_\text{\tiny{outer}} \left( \bigoplus_{i \in N} \phi_\text{\tiny{inner}} \left( q_i\scaleobj{0.8}{(\boldsymbol{\tau}_{\mathcal{N}_i},u_i)} \right) \right)
    \label{deepsets}
\end{equation}
		
where $\phi_\text{\tiny{inner}}$ and $\phi_\text{\tiny{outer}}$ denote multi layer perceptrons. In practice, we substitute the sum operation in Deep Sets for Powermean \cite{pnorm}, supplying an inductive bias which simplifies the representation of nonlinear aggregation functions like min and max.  Because interactions between agents are accounted for in the model, the QGNN mixer does not require a hypernetwork like QMIX \cite{QMIX}. The IGM condition \eqref{eq:IGM} in QGNN is addressed in the same manner as QMIX: the weights of $\phi_\text{\tiny{inner}}$ and $\phi_\text{\tiny{outer}}$ are constrained to be positive (only the matrix multiplication weights need to be positive to ensure monotonicity, so the biases are unrestricted). This enforces the monotonicity constraint $\forall i, \, \frac{\partial Q}{\partial q_i} \geq 0$, which satisfies the IGM condition.

\subsection{Training}

The underlying reinforcement learning algorithm used to train the network is a DQN \cite{DQN}. The network parameters $\theta$ are updated by minimising the squared TD error $\mathcal{L}(\theta) = \mathbb{E} \left[ \left( V_\mathrm{target} - Q(\boldsymbol{\tau}_{t}, \textbf{u}_t, \theta) \right)^2 \right]$, where $V_\mathrm{target} = r_t + \gamma \argmax_{\textbf{u}'}Q(\boldsymbol{\tau}_{t+1}, \textbf{u}',\theta^{-})$. The parameters of the target network $\theta^{-}$ are periodically updated with $\theta$.


\section{Experiments}
\label{sec:experiments}

\begin{figure*}[t]

    \begin{subfigure}[t]{0.45\textwidth}
    	\centering
        \includegraphics[width=0.99\linewidth]{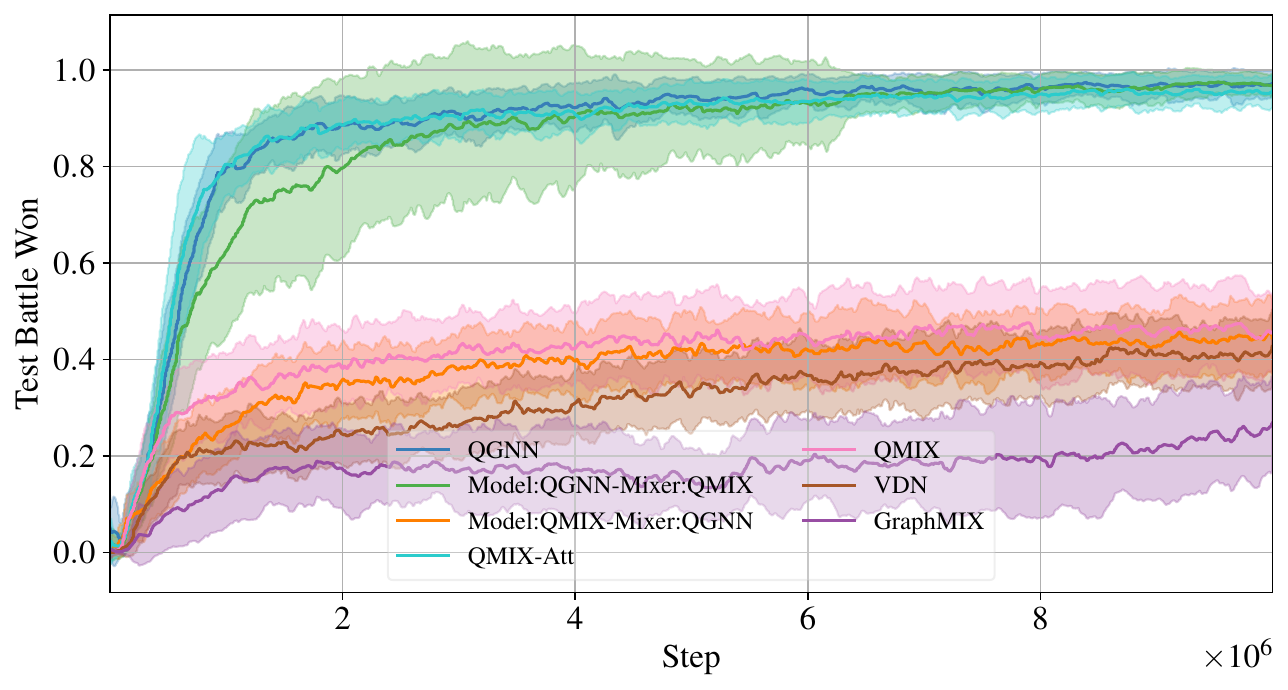}
    	\caption{Mean percentage of battles won in the Starcraft 1o\_10b\_vs\_1r environment.}
    	\label{fig:starcraft}
    \end{subfigure}
    \hfill
    \begin{subfigure}[t]{0.45\textwidth}
    	\centering
        \includegraphics[width=0.99\linewidth]{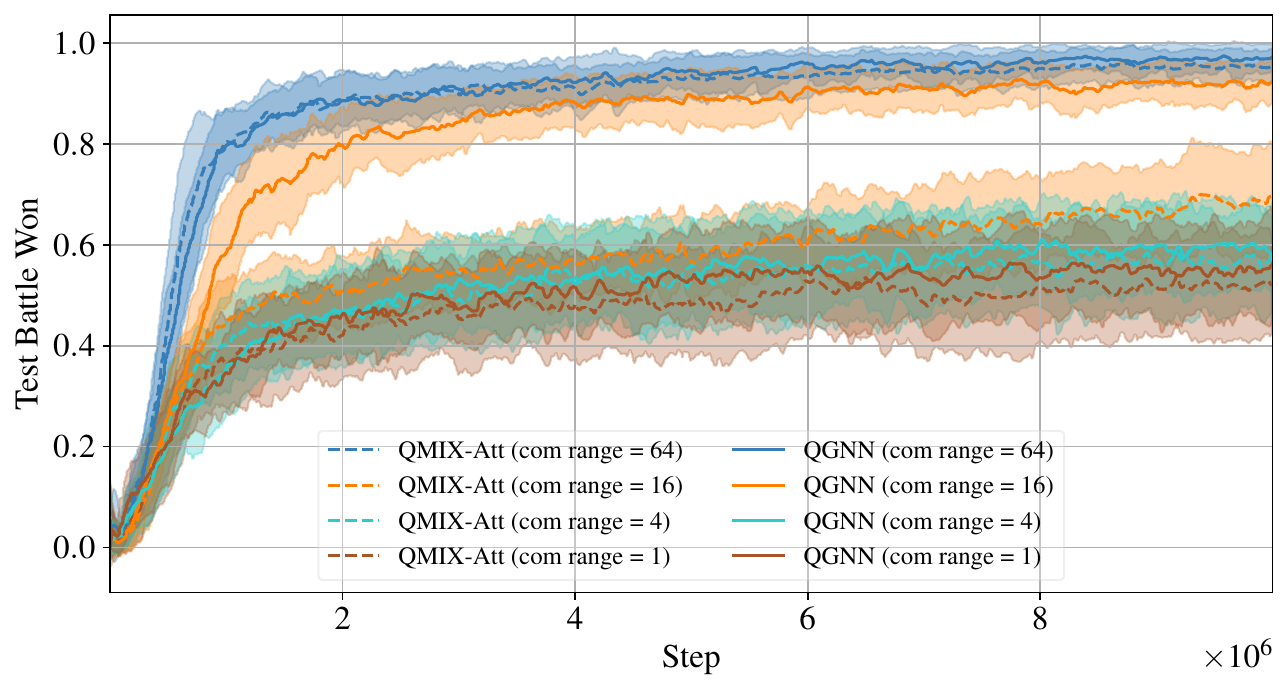}
    	\caption{An analysis of the effect of communication range on the performance (mean battle won) of QGNN and QMIX-Att-Graph in Starcraft 1o\_10b\_vs\_1r.}
    	\label{fig:starcraft_com}
    \end{subfigure}
    \hfill
    \begin{subfigure}[t]{0.45\textwidth}
    	\centering
        \includegraphics[width=0.99\linewidth]{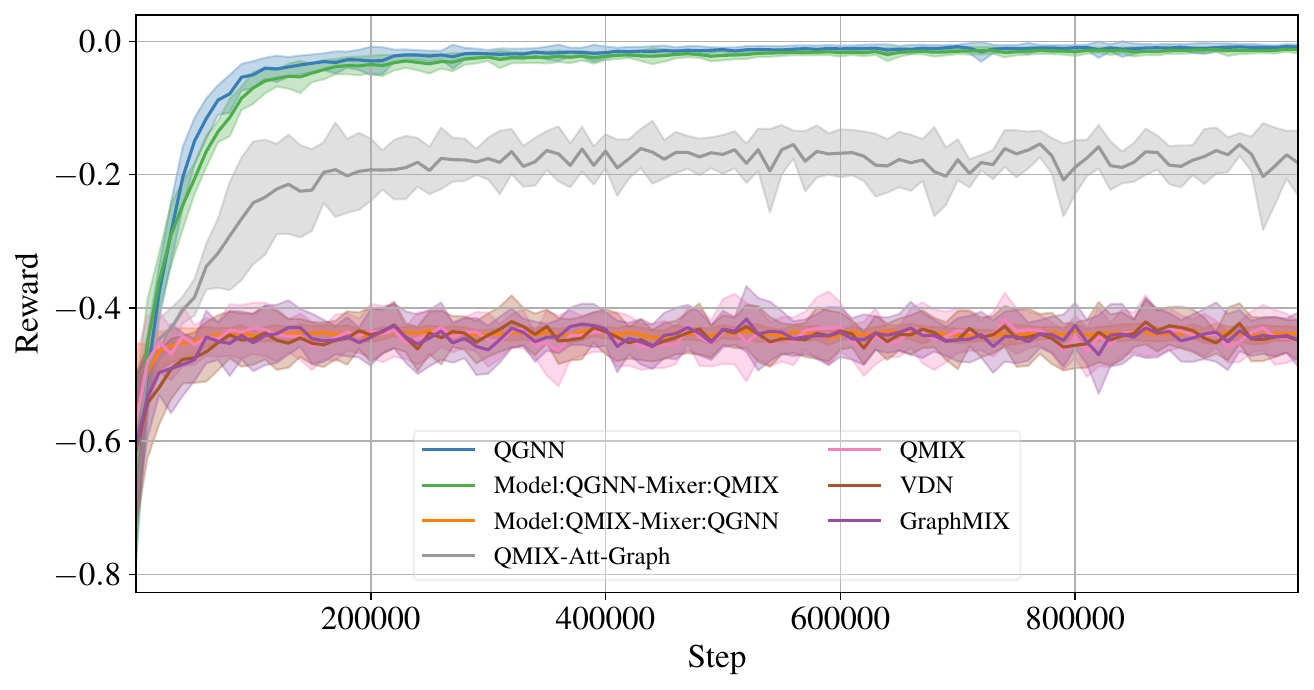}
    	\caption{Mean reward in the Estimate Game.}
    	\label{fig:estimate}
    \end{subfigure}
    \hfill
    \begin{subfigure}[t]{0.45\textwidth}
    	\centering
        \includegraphics[width=0.99\linewidth]{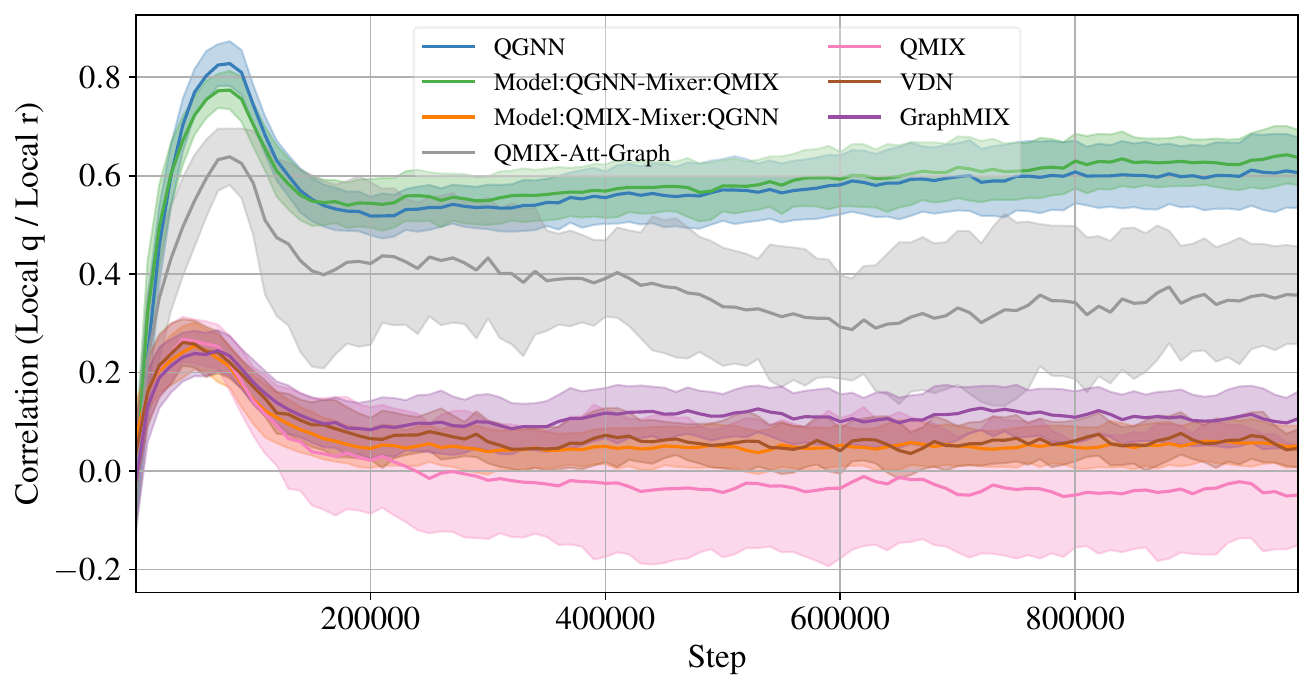}
    	\caption{An analysis of the accuracy of the local values (correlation coefficient with respect to \say{ground truth} local rewards) in the Estimate Game.}
    	\label{fig:estimate_corr}
    \end{subfigure}
    \hfill
    \begin{subfigure}[t]{0.45\textwidth}
    	\centering
        \includegraphics[width=0.99\linewidth]{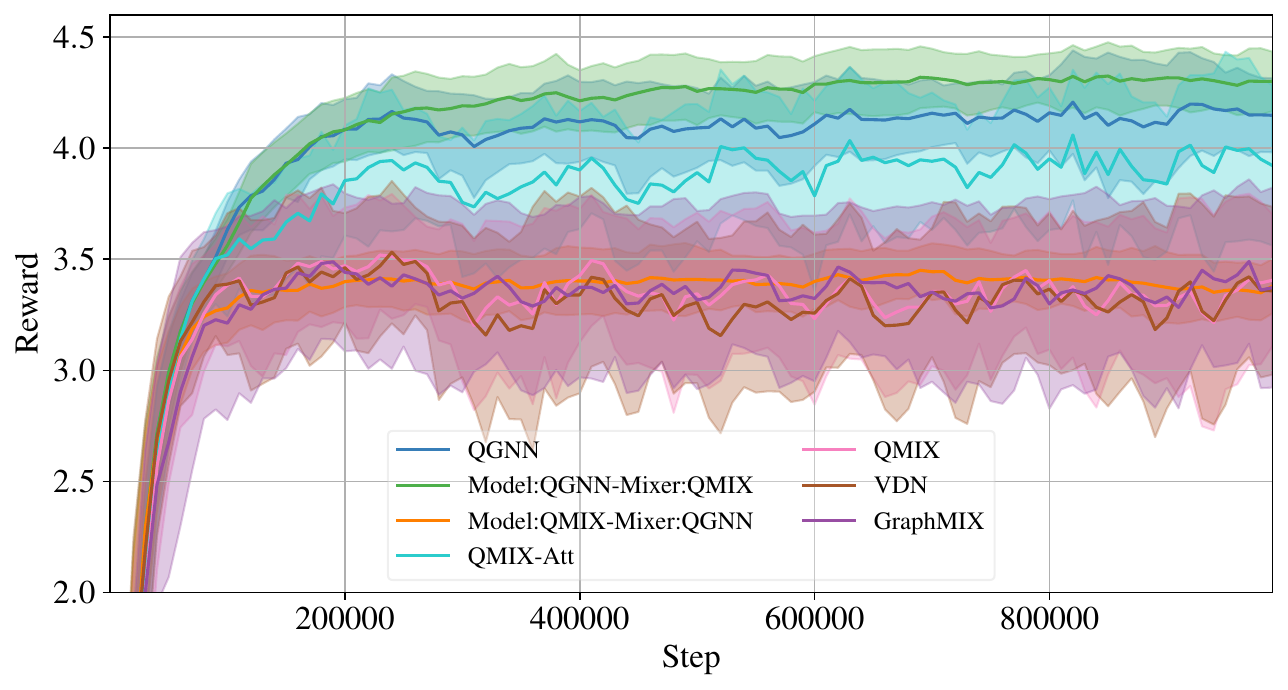}
    	\caption{Mean reward in the Coalition Structure Generation Problem.}
    	\label{fig:set}
    \end{subfigure}
    \hfill
    \begin{subfigure}[t]{0.45\textwidth}
    	\centering
        \includegraphics[width=0.99\linewidth]{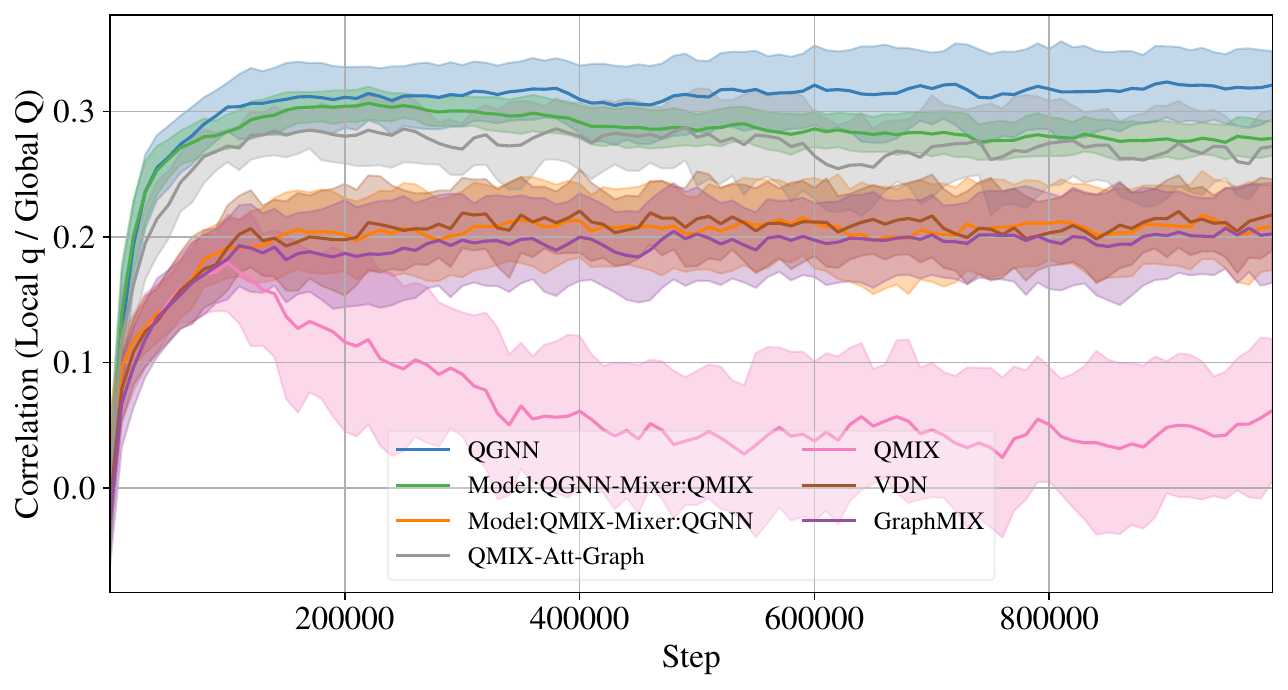}
    	\caption{Correlation between the predicted local values and the global reward in the Estimate Game.}
    	\label{fig:estimate_global_corr}
     \hfill
    \end{subfigure}

    \caption{Experimental results in the Starcraft, Estimate Game, and Coalition Structure Generation environments. The shaded regions represent the min and max across all runs. }
    \label{fig:results}
\end{figure*}

In this section, we introduce our experimental setup and analyse the results (see Appendix \ref{appendix:training} for training details). We evaluate on the following environments:

\begin{itemize}[leftmargin=0.4cm,noitemsep,topsep=0pt]
    \item \textbf{Starcraft II}. A game commonly used for benchmarking MARL algorithms.
	\item \textbf{Estimate Game}. A custom environment that explicitly models inter-agent dependencies. 
	\item \textbf{Coalition Structure Generation}. A foundational NP-hard problem which can be tackled from a multi-agent perspective.
\end{itemize}

We evaluate QGNN against recent state-of-the-art value factorisation methods: QMIX-Att \cite{AI-QMIX} (the only other value factorisation method which uses communication), GraphMIX \cite{GraphMIX} (a method which uses a GNN in the mixer), Fine-tuned QMIX \cite{QMIX} (an implementation of QMIX in PyMARL2 \cite{Pymarl2} which has been fine-tuned to achieve state of the art performance), and VDN \cite{VDN} (the original value factorisation method). In order to account for an unfair advantage due to utilising the graph structure of our environments, we also introduce QMIX-Att-Graph, a variant of QMIX-Att which operates over the same communication graph as QGNN. 

Our experiments also include hybrid models, which provide an ablation analysis over the main components in our architecture. These hybrid models are generated by replacing the QGNN model or mixer with the analagous component in QMIX. In the plot legends, hybrid architectures are described by the naming convention: Model:[model type]--Mixer:[mixer type]. 

In our results, we report the ``maximum'' reward, battle won percentage, or correlation coefficient, which is calculated by taking the mean across all runs and the maximum across all timesteps. All experiments consist of a minimum of 10 runs.


\subsection{Starcraft II}
\label{section:starcraft}

\begin{figure}
    \centering
    \includegraphics[width=0.6\linewidth]{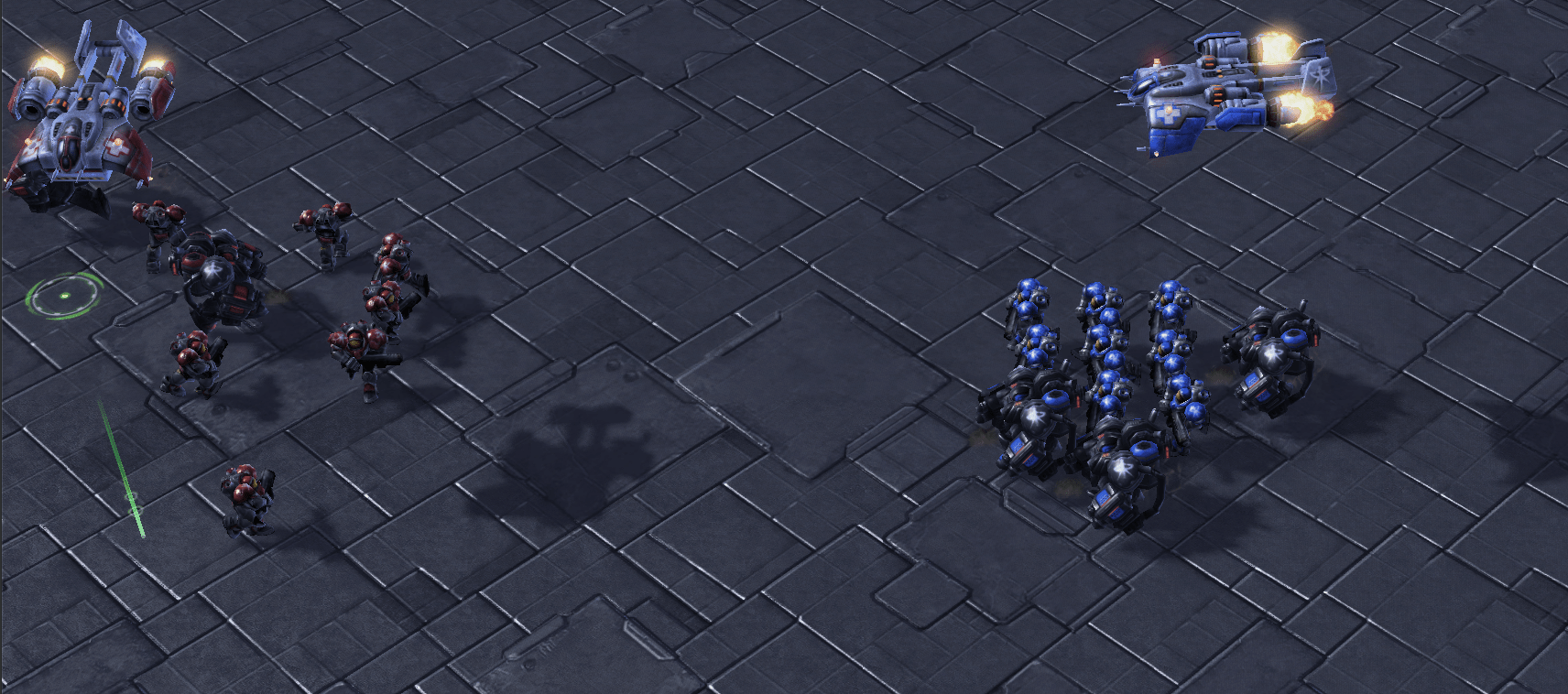}
    \caption{The SMAC Environment. In the Starcraft Multi-Agent challenge, the task is to apply micromanagement to every unit in team, with the goal of destroying the opposing AI-controlled team. }
    \label{fig:smac}
\end{figure}

\textbf{Environment.} Starcraft has established itself as the standard method of benchmarking credit assignment algorithms. It is a competitive simulation environment in which two teams of individually controlled agents fight to the death. The task is fully cooperative---the entire team receives a positive reward for a victory, and negative for defeat. Consequently, the problem lends itself to a credit assignment approach.

In our experiments, we use the SMAC environment \cite{SMAC}. In SMAC, the observation consists of the distance, relative position, health/shield, unit type, and last action of allies and enemies within a limited field of view (although last action is only available for allies). The agents can observe terrain features. The action space consists of four movement directions, an attack action for each enemy, and no-op. The global reward function is shaped---the system receives a small reward for damaging or killing an enemy, and a large reward for winning. In our experiments we examine the 1o\_10b\_vs\_1r because it is one of the most difficult environments, and it seems to require more inter-agent coordination than other environments (according to the results in PyMARL2 \cite{Pymarl2}).

\textbf{Results.} Performance in Starcraft falls into two distinct categories (Fig. \ref{fig:starcraft}). QGNN and QMIX-Att solve the problem, achieving maximum battle won percentages of $0.98$ and $0.96$, respectively. In contrast, the non-communication-based methods (GraphMIX, QMIX, VDN) collectively achieve a maximum battle won percentage of $0.49$.

In the Starcraft environment, we also perform an ablation over communication range for both QGNN and QMIX-Att-Graph (Fig. \ref{fig:starcraft_com}). Our first finding from this experiment is that as the communication range approaches zero, the performance of QGNN approaches that of QMIX. While this behaviour is expected due to the similarity between the QMIX model and the QGNN model on a communication graph with with no edges, it does show that performance can be improved even with a subset of the edges from the \say{true} graph. Another interesting finding from this experiment is the degree to which QGNN outperforms QMIX-Att-Graph with equivalent communication ranges. Of course, part of this difference in performance can be attributed to the fact that QGNN contains a multi-layer GNN, allowing it to increase its receptive field. However, this does not entirely explain the discrepancy. The performance of QGNN with a communication range of 4 is comparable to the performance of QMIX-Att-Graph with a communication range of 16, even though the theoretical maximum receptive field of QGNN's 2-layer GNN is 8. Furthermore, QGNN with a communication range of 16 nearly matches the performance of the fully connected version with a maximum battle won percentage of $0.93$, while QMIX-Att-Graph with the same connectivity only achieves a percentage of $0.70$, which is marginally better than QMIX.

\subsection{Estimate Game}

\begin{figure}
    \centering
    \includegraphics[width=0.5\linewidth]{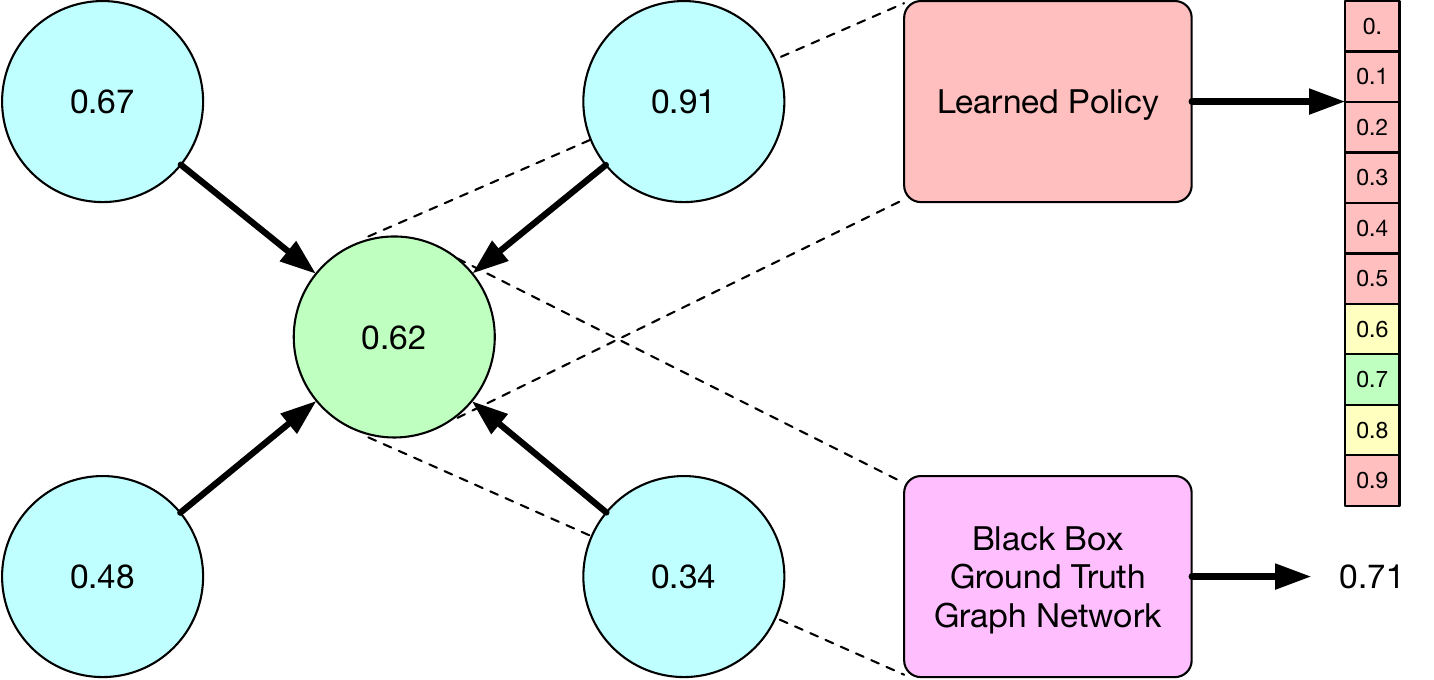}
    \caption{The Estimate Game Environment. The environment includes a black box graph network, which deterministically produces an output as a function of the state of each node and its neighbours. The goal is for the learned policy to estimate the output of the ground truth function by selecting the corresponding action. QGNN (and the QMIX-Att-Graph baseline) use the same connectivity graph as the ground truth graph network. In this diagram, the nodes are represented as circles, and the edges as arrows.}
    \label{fig:est}
\end{figure}

\textbf{Environment.} While most sufficiently complex multi-agent environments incorporate inter-agent dependencies, it is not always clear how those interactions manifest themselves in the reward function. 

In order to analyse the capacity of MARL methods to model these interactions, we introduce the Estimate Game: a multi-agent environment which explicitly models interdependent local reward functions. In the game, each of the $n=8$ agents receives a local state $x_i \in [0,1]$ as an observation. The agents are also assigned a random graph connectivity with an edge density of 20\%. Then, a black-box function of each agent's local state $x_i$ and neighbours' states $x_j \in \mathcal{N}_i$ is used to produce a hidden state for each agent $y_i \in [0,1]$. In practice, this black-box function is implemented as \inlineequation[eq:estimate]{$y_i = 2 \left(0.3 (x_i-0.5) + 0.7 \frac{1}{\lvert \mathcal{N}_i \rvert} \sum_{j \in \mathcal{N}_i} (x_j-0.5) \right) + 0.5$}. In effect, this formula mixes the local and neighbouring states with weighting $0.3$ and $0.7$ respectively, and doubles the variance of the resulting distribution. The goal for each agent is to guess this hidden state $y_i$. Each agent has $10$ actions, corresponding to guesses splitting the domain into $10$ intervals $[0,0.1], \dots, [0.9,1]$. The local reward for each agent is proportional to the error with respect to the closest value in the predicted interval. For example, if the true value is $0.36$, then guessing the interval $[0.3,0.4]$ will earn a reward of $0$, while guessing the interval $[0.5,0.6]$ will incur a reward of $-0.14$. Finally, the global reward is calculated by taking the minimum of all of the local rewards.

The Estimate Game is a useful benchmark because it provably requires coordination to solve, as shown by the explicit dependency on non-local information in eq. \eqref{eq:estimate}. This provides complimentary information to environments like Starcraft, in which the optimal policy is assumed to require a complex coordination strategy, but is ultimately unknown. While many methods can now solve Starcraft, the Estimate Game serves as an effective stress test to differentiate those methods.

\textbf{Results.} The results show that QGNN outperforms all of the baselines in the Estimate Game (Fig. \ref{fig:estimate}). It is the only method which is able to solve the problem, achieving the maximum possible reward of $0$. The next best baseline method is our modified version of QMIX-Att which operates over graphs. However, its single-layer attention mechanism lacks the representational complexity to predict the correct action.

While the Estimate Game is just a constructed example, it does provide some insight about the limitations of existing methods. QMIX-Att can utilise communication to outperform methods like GraphMIX, QMIX, and VDN, but it lacks the capacity to model complex interactions between agents.

Since the Estimate Game explicitly models the reward as a function of the joint state and action, it possesses \say{ground truth} local rewards. This provides a unique opportunity to examine the ability of each model to perform credit assignment. The IGM condition does not pose any constraints on the relative scale of the rewards for different agents as a result of their actions. Consequently, the act of ascribing values to agent's individual contributions must be learned implicitly.

Since the scale and shift of the implicitly learned local q-values is undefined, we use the correlation coefficient between the q-values and \say{ground truth} local rewards as a metric for accuracy (Fig. \ref{fig:estimate_corr}). The non-communication-based methods achieve a maximum correlation coefficient of $0.27$, while QMIX-Att-Graph reaches $0.64$ and QMIX-Att reaches $0.28$. In contrast, QGNN attains a maximum correlation coefficient of $0.83$. 

The biggest insight about this experiment can be gleaned from the difference between the local value vs \say{ground truth} local reward correlation in Fig. \ref{fig:estimate_corr} and the local value vs global reward correlation in Fig. \ref{fig:estimate_global_corr}. Even though they were trained on the global reward, some methods are able to achieve a \textit{higher} correlation with the local rewards. This implicitly learned relationship is an emergent behaviour due to the manner in which the models compress information. Since QGNN possesses the most representational complexity, it is able to learn the most accurate local q-values.

\subsection{Coalition Structure Generation}

\begin{figure}
    \centering
    \includegraphics[width=0.99\linewidth]{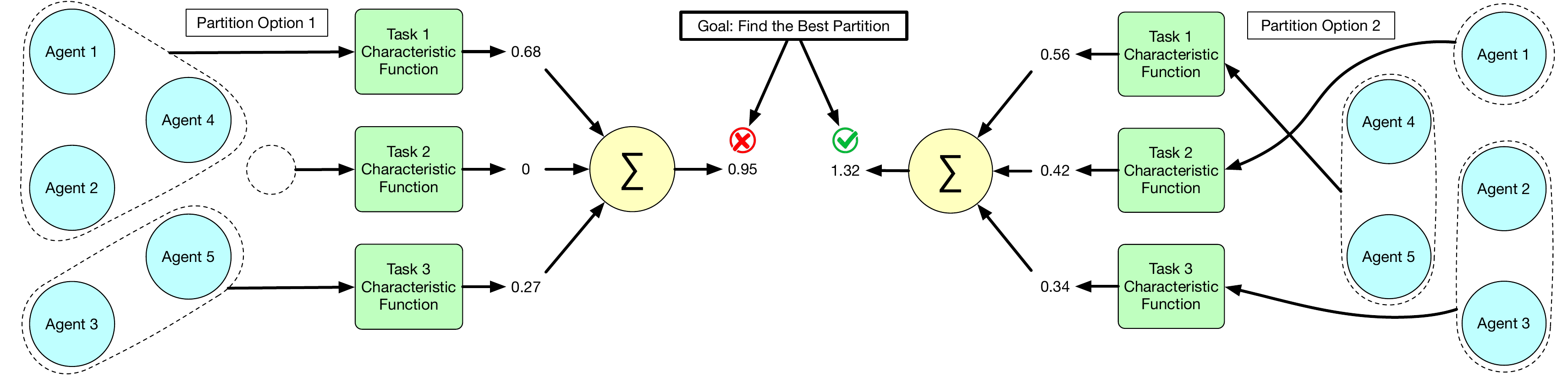}
    \caption{The Coalition Structure Generation Environment. In this problem, $n$ agents must be assigned to $m$ tasks. For each assignment, the entire team receives some reward. This reward is calculated is calculated as the sum of the outputs of the (nonlinear) characteristic functions, each of which computes a value based on the set of agents assigned to it. The goal of the task is to find the partition which maximises the team's reward. Since the individual value functions are interdependent, QGNN uses a fully connected graph in this scenario---there are edges (not shown in the figure) between every pair of agents.  }
    \label{fig:csg}
\end{figure}

\textbf{Environment.} The Coalition Structure Generation problem, also known as the Complete Set Partitioning problem, is of great interest due to its diverse selection of applications, including combinatorial auctions, sensor networks, and characteristic function games \cite{CSP}. However, the problem is NP-hard, so there is a limit to the effectiveness of first-principles solutions \cite{CSP-NPhard}. In this experiment, we attempt to approximate the optimal solution with a learning-based method. In contrast with most experiments in related work, this demonstrates the performance of credit assignment on a real-world problem.

The Coalition Structure Generation problem concerns the assignment of a set of $n$ agents $N = \{1..n\}$ to $m$ tasks. Each task is associated with a characteristic function $\nu_j \in V$ which maps a subset of agents to a value $\nu_j : 2^N \rightarrow \mathbb{R}$. A solution is characterised by a set $\mathcal{S} = \{S_1,\dots,S_m\}$ of disjoint subsets of agents mapped to each task, such that $\bigcup_{S_j \in \mathcal{S}} S_j = N$. The value of a solution $\mathcal{S}$ is given by: $J(\mathcal{S}) = \sum_{j=1}^m \nu_j(S_j)$. The objective is to find $\mathcal{S}^* = \argmax_{\mathcal{S}} J(\mathcal{S})$. Since all agents must coordinate their assignments simultaneously, the communication graph in this problem is defined to be fully connected.

In our experiments, we define the characteristic function by generating a random $n \times m$ matrix ($n=16$, $m=4$) drawn from a gamma distribution ($\alpha=2$, $\beta=1$). This matrix represents the rewards for each agent with respect to each task. The characteristic functions $\nu_j$ define the value of a coalition as the mean of the agent-task assignment values. However, if a coalition is empty, then it receives a reward of $-10$. In accordance with the problem formalisation, the global reward is defined as the sum of the values of the coalitions for each task.

\textbf{Results.} The results in this task mirror those of the Estimate Game (Fig. \ref{fig:set}). The QGNN model outperforms all of the baselines, with a maximum reward of $4.21$ for the default model and $4.33$ for the QGNN model with the QMIX mixer. As expected, the communication-based baseline QMIX-Att achieves the second best performance, with a maximum reward of $4.06$. The three non-communication-based methods all plateau at the same level, which likely represents the optimal performance of a locally greedy policy. Collectively, they achieve a maximum reward of $3.53$.


\section{Discussion}
\label{sec:discussion}

QGNN outperforms all of the baselines (to varying degrees) across all experiments (Fig. \ref{fig:starcraft}, \ref{fig:estimate}, \ref{fig:set}). We evaluate against both QMIX-Att and QMIX-Att-Graph, showing that QGNN's superior performance is not only a consequence of the sparsity of the graph over which it operates, but also the representational complexity of the architecture itself (Fig. \ref{fig:estimate}). 

In the Starcraft experiment, we also show that progressively pruning edges in the communication graph indeed causes the performance of QGNN to approach that of QMIX (Fig. \ref{fig:starcraft}). This experiment is useful for explainability with respect to the impact of communication, and it allows users to incrementally transition from a non-communicative model (\textit{e.g.} QMIX), to one that can use communication to different degrees. 

In all of the experiments, we perform an ablation analysis over each component of QGNN. The results show that swapping the QGNN model for the QMIX or QMIX-Att model results in a significant boost in performance (Fig. \ref{fig:starcraft}, \ref{fig:estimate}, \ref{fig:set}). On the other hand, swapping the QMIX mixer with the QGNN mixer has very little effect, supporting our claim that additional complexity is more important in the model than the mixer. Our mixer has a simpler formulation with no hypernetwork, contains far fewer parameters, and does not scale in size with the number of agents (as QMIX does). In a system with 8 agents, the QGNN mixer has 193 parameters, and the QMIX mixer has 20,481.

In addition to comparing the overall performance of QGNN and the baselines, we also examine the actual local q-value predictions. We show that, in contrast with the baselines, QGNN can implicitly learn the \say{correct} local values (Fig. \ref{fig:estimate}). This makes a strong statement, as it is not necessary to perform credit assignment in order to satisfy the IGM condition. Any communication-based method with sufficient representational complexity could simply predict the global Q-value for each agent, and that would represent a valid factorisation. However, many global value functions exhibit some degree of independence which allow them to be represented as a composition of components which are conditioned on non-global information (for example, in environments with a spatial component, distant agents often do not affect each other). In such scenarios, our method's GNN-based architecture naturally factorises the global value function, implicitly performing credit assignment.

Collectively, these results show the effectiveness of QGNN in complex problems---its multi-layer communication-based model allows it to solve problems which require information sharing (like the Estimate Game) and problems with non-monotonic value functions (like the Coalition Structure Generation Problem). In Appendix \ref{appendix:limitations}, we discuss some of the limitations of our work.


\section{Conclusion}

This paper presented QGNN, the first value factorisation method with a GNN-based model. Our method outperformed all of the baselines in all of the experiments, which included a popular baseline, a custom problem to serve as a \say{stress test}, and a foundational NP-hard problem. In addition to its performance benefits, QGNN provides several advantages due to its graph structure: it can model constraints like communication range in real world systems, it can encode graph topological priors with information about interdependencies, and it allows users to incrementally add communication to a base non-communicative model. Finally, a key contribution of this paper is the insight that \textbf{a)} a model conditioned on non-local information with sufficient representational complexity is needed to represent a global value function with interdependence (regardless of the mixer which is used), \textbf{b)} even if each agent can predict the global value, a model with inductive biases can incentivise factorisation, and \textbf{c)} GNNs satisfy both of these criteria. In this work, we demonstrate that GNNs implicitly factorise graph-level predictions, and thus can be used as effective value factorisation methods with little modification.


\section*{Acknowledgements}

Ryan Kortvelesy is supported by Nokia Bell Labs through their donation for the Centre of Mobile, Wearable Systems and Augmented Intelligence to the University of Cambridge. A. Prorok acknowledges funding through ERC Project 949940 (gAIa).


\newpage

\begin{appendices}

\section{Non-Monotonic Value Functions}
\label{appendix:nonmonotonic}

Functions where the added value of any agent is dependent on the actions taken by other agents can be non-monotonic. Consider this example of a cooperative matrix game where the actions of both agents $\mathbf{u} = [u_0; u_1]$ are taken simultaneously (Table \ref{MatrixGame}).

\begin{table}[h]
\begin{center}
\begin{tabu}{|[1pt]c|[1pt]c|c|[1pt]}
     \tabucline[1pt]{}
     \backslashbox{$u_0$}{$u_1$} & 0 & 1 \\
     \tabucline[1pt]{}
     $0$ & $1.0$ & $0.0$ \\
     \hline
     $1$ & $0.0$ & $1.0$ \\
     \tabucline[1pt]{}
\end{tabu}
\caption{\small{The payoffs for each joint action in a matrix game.}}
\label{MatrixGame}
\end{center}
\end{table}

When we fit a continuous value function $Q$ to this payoff matrix, notice that the component of the gradient of $Q$ due to agent 0's action (which corresponds to the unit vector $[1,0]$) is $\frac{\partial Q}{\partial \mathbf{u}} \vert_{\mathbf{u}=[0,0]} \bullet [1,0] < 0$ evaluated at $[0,0]$, but it is $\frac{\partial Q}{\partial \mathbf{u}} \vert_{\mathbf{u}=[0,1]} \bullet [1,0] > 0$ evaluated at $[0,1]$. In other words, increasing the value of $u_0$ will decrease the global reward if $u_1=0$, but it will increase the global reward if $u_1=1$. Furthermore, remember that $q_i$ is only a function of the local action $u_i$, so $\frac{\partial q_0}{\partial u_0} \vert_{u_1=0} = \frac{\partial q_0}{\partial u_0} \vert_{u_1=1}$. Therefore, no matter what function $q_i$ is, either $\left[ \left(\frac{\partial Q}{\partial \mathbf{u}} \bullet [1,0] \right) \frac{\partial u_0}{\partial q_0} \right] \vert_{\mathbf{u}=[0,0]} = \frac{\partial Q}{\partial q_0} \vert_{\mathbf{u}=[0,0]} < 0$ or $\left[ \left(\frac{\partial Q}{\partial \mathbf{u}} \bullet [1,0] \right) \frac{\partial u_0}{\partial q_0} \right] \vert_{\mathbf{u}=[0,1]} = \frac{\partial Q}{\partial q_0} \vert_{\mathbf{u}=[0,1]} < 0$. The condition for monotonicity is that $\frac{\partial Q}{\partial q_i} \geq 0 \; \forall i \in \{1..N\}$, so in this example $Q$ is not monotonic. 

Note that if non-local information can be used, then the value of an action $u_0$ can depend on the value of $u_1$, such that $\frac{\partial q_0}{\partial u_0} \vert_{u_1=0} \neq \frac{\partial q_0}{\partial u_0} \vert_{u_1=1}$. Consequently, it would be possible for the monotonicity constraint to be satisfied.

\section{Shapley Values}
\label{appendix:shapley}

To demonstrate that a state-agnostic mixer can prove effective if the model possesses sufficient representational complexity, consider the calculation of Shapley values \cite{Shapley}. A Shapley value is a method for distributing reward in a cooperative, coalitional game. Given a set $N$ of $n$ agents, $v: 2^N \rightarrow \mathbb{R}$ is a value function which maps a coalition $S \subseteq N$ to a reward. The Shapley value $\psi_i(v)$ for agent $i$ is defined:

	$$ \psi_i(v) = \frac{|S|!\,(n-|S|-1)!}{n!} \left( v(S \cup \{i\}) - v(S) \right)$$
	
	Furthermore, the value of the grand coalition $v(N)$ equals the sum of the Shapley values for each agent:
	
	$$ v(N) = \sum_{i \in N} \psi_i(v)$$

This first-principles derivation lends some insight into the most appropriate architecture for credit assignment. Firstly, note that the ``mixing'' function is simply a summation. Secondly, note that the local Shapley value $\psi_i(v)$ is conditioned upon the global value function $v$, which means that it requires global state information. This illustrates that a simple, state-agnostic mixer is sufficient for representing local values. It also shows that non-local information (which can be obtained with communication) is required to predict local values.

\section{Kolmogorov-Arnold Representation Theorem}
\label{appendix:kolmogorov}
The Kolmogorov-Arnold representation theorem states that any multivariate function can be represented as a composition of univariate functions and summations \cite{Kolmogorov}:
	
\begin{equation}
    f(x_1, \dots, x_n) = \sum_{q=0}^{2n} \Phi_q \left( \sum_{p=1}^{n} \phi_{p,q}(x_p) \right) .
\end{equation}
		
Since its introduction, it has been shown that the number of outer functions $\Phi_q$ can be reduced to a single function $\Phi$ without a loss in representational complexity \cite{KolmogorovSimplification}. Furthermore, if the function space is constrained to symmetric (permutation invariant) functions, then the set of inner functions $\phi_{p,q}$ can be replaced with a single function $\phi$. Recently, Deep Sets has demonstrated the effectiveness of this modified formulation for encoding sets \cite{DeepSets}.

\section{Training Details}
\label{appendix:training}
All experiments are conducted in PyMARL2 \cite{Pymarl2} (Apache-2.0 License), a library with baseline multi-agent reinforcement learning algorithms which have been fine-tuned to produce significantly better results than originally reported. To avoid giving QGNN an unfair advantage with hyperparameter tuning, we use the same hyperparameters as QMIX:

\begin{itemize}
    \item Experience is collected into a buffer of size $5000$, and training occurs with batch sizes of $128$.
    \item Exploration is handled by an $\epsilon$-greedy policy, where $\epsilon$ is annealed from $1.0$ to $0.05$ over $10^5$ timesteps.
    \item During training, value function updates are calculated with T($\lambda$), where $\lambda=0.6$.
    \item The learning rate is set to $10^{-3}$.
    \item Every $200$ steps, the target network is updated.
    \item The model maps the input dimension to a hidden size of $64$.
\end{itemize}

Instead of performing an exhaustive grid search to find the optimal set of hyperparameters for each method on each experiment (which we do not have the computational resources for), we use each algorithm's default hyperparameters on all experiments. Note that the baseline algorithms' default hyperparameters are tuned to maximise performance on the starcraft environment.

All of the experiments in this paper were conducted on a GeForce RTX 2080 Ti 11G GPU. The time taken to run each experiment varied, but for Starcraft it could take up to 30 hours to perform one run of a single method. It took approximately one month to run all of the experiments.

\section{Limitations}
\label{appendix:limitations}
QGNN specifically focuses on the value factorisation problem. It does not address other challenges in reinforcement learning, like learning in environments with very sparse rewards. One limitation of QGNN stems from the fact that it is built upon the DQN learning scheme, and is therefore constrained to discrete actions. Consequently, QGNN would not be well-suited to problems with continuous action spaces (which encompasses most real-world robotics problems). Another limitation of DQN is the manner in which it computes q-values for each action---instead of actions being treated as inputs, the network produces a separate output for each possible action. This means that the actions of other agents must be inferred based on their states. While it is easy to infer these actions with homogeneous, deterministic policies, it is not possible to take other agents' actions into account with a stochastic policy.

Another limitation of QGNN is rooted in its use of communication. In problems which can be run on a single machine, such as video games (such as Starcraft) and computational problems (such the coalition structure generation problem), the act of \say{communication} between agents does not require any additional considerations. However, if QGNN is executed on real-world robots, then there are several problems which could arise which we have not addressed in this work, such as communication delays, dropped messages, and adversarial agents.

While there is no guarantee that these results will generalise to other environments, we note that there is no selection bias in the environments in this paper. That is, the environments were not tweaked to influence the results of any individual method, and we did not discard any environments due to unsatisfactory results. For the sake of transparency, we did run simpler Starcraft environments like 3m, 3s5z, and 8m. However, we found that every single method solved these environments, so they did not constitute a suitable benchmark. Consequently, we switched to a more difficult environment to produce more interesting results.

\end{appendices}

\newpage
\section*{Declarations}

\subsection*{Ethical Approval}
Not Applicable.

\subsection{Competing Interests}
Apart from the funding source (see the Funding section), the authors have no competing interests to declare which are relevant to this article.

\subsection{Authors' Contributions}
This article was written by Ryan Kortvelesy and supervised by Amanda Prorok.

\subsection{Funding}
This work was supported by Nokia Bell Labs and the ERC (Project 949940).

\subsection{Availability of Data and Materials}
All of the code needed to reproduce the results in this paper can be found at: \url{https://github.com/Acciorocketships/pymarl2}


\newpage

\bibliographystyle{unsrt}
\bibliography{main}

\begin{thebibliography}{10}

\bibitem{hyldmar_Fleet_2019}
Nicholas Hyldmar, Yijun He, and Amanda Prorok.
\newblock A {Fleet} of {Miniature} {Cars} for {Experiments} in {Cooperative}
  {Driving}.
\newblock {\em IEEE International Conference Robotics and Automation (ICRA)},
  2019.

\bibitem{SensorNets}
M.~Rabbat and R.~Nowak.
\newblock Distributed optimization in sensor networks.
\newblock In {\em Third International Symposium on Information Processing in
  Sensor Networks, 2004. IPSN 2004}, pages 20--27, 2004.

\bibitem{ModControl}
Wenlong Huang, Igor Mordatch, and Deepak Pathak.
\newblock One policy to control them all: Shared modular policies for
  agent-agnostic control.
\newblock In {\em International Conference on Machine Learning}, pages
  4455--4464. PMLR, 2020.

\bibitem{MultiagentRewards}
Adrian~K Agogino and Kagan Tumer.
\newblock Analyzing and visualizing multiagent rewards in dynamic and
  stochastic domains.
\newblock {\em Autonomous Agents and Multi-Agent Systems}, 17(2):320--338,
  2008.

\bibitem{POMDP}
Frans~A Oliehoek, Christopher Amato, et~al.
\newblock {\em A concise introduction to decentralized POMDPs}, volume~1.
\newblock Springer, 2016.

\bibitem{IQL}
Ming Tan.
\newblock Multi-agent reinforcement learning: Independent vs. cooperative
  agents.
\newblock In {\em Proceedings of the tenth international conference on machine
  learning}, pages 330--337, 1993.

\bibitem{VDN}
Peter Sunehag, Guy Lever, Audrunas Gruslys, Wojciech~Marian Czarnecki, Vinicius
  Zambaldi, Max Jaderberg, Marc Lanctot, Nicolas Sonnerat, Joel~Z Leibo, Karl
  Tuyls, et~al.
\newblock Value-decomposition networks for cooperative multi-agent learning
  based on team reward.
\newblock In {\em Proceedings of the 17th International Conference on
  Autonomous Agents and MultiAgent Systems}, pages 2085--2087, 2018.

\bibitem{QMIX}
Tabish Rashid, Mikayel Samvelyan, Christian Schroeder, Gregory Farquhar, Jakob
  Foerster, and Shimon Whiteson.
\newblock Qmix: Monotonic value function factorisation for deep multi-agent
  reinforcement learning.
\newblock In {\em International Conference on Machine Learning}, pages
  4295--4304. PMLR, 2018.

\bibitem{QTRAN}
Kyunghwan Son, Daewoo Kim, Wan~Ju Kang, David~Earl Hostallero, and Yung Yi.
\newblock Qtran: Learning to factorize with transformation for cooperative
  multi-agent reinforcement learning.
\newblock In {\em International Conference on Machine Learning}, pages
  5887--5896. PMLR, 2019.

\bibitem{GraphMIX}
Navid Naderializadeh, Fan~H Hung, Sean Soleyman, and Deepak Khosla.
\newblock Graph convolutional value decomposition in multi-agent reinforcement
  learning.
\newblock {\em arXiv preprint arXiv:2010.04740}, 2020.

\bibitem{qplex}
Jianhao Wang, Zhizhou Ren, Terry Liu, Yang Yu, and Chongjie Zhang.
\newblock Qplex: Duplex dueling multi-agent q-learning.
\newblock {\em arXiv preprint arXiv:2008.01062}, 2020.

\bibitem{qrelation}
Siqi Shen, Jun Liu, Mengwei Qiu, Weiquan Liu, Cheng Wang, Yongquan Fu, Qinglin
  Wang, and Peng Qiao.
\newblock Qrelation: an agent relation-based approach for multi-agent
  reinforcement learning value function factorization.
\newblock In {\em ICASSP 2022 - 2022 IEEE International Conference on
  Acoustics, Speech and Signal Processing (ICASSP)}, pages 4108--4112, 2022.

\bibitem{raca}
Hao Chen, Guangkai Yang, Junge Zhang, Qiyue Yin, and Kaiqi Huang.
\newblock Raca: Relation-aware credit assignment for ad-hoc cooperation in
  multi-agent deep reinforcement learning.
\newblock In {\em 2022 International Joint Conference on Neural Networks
  (IJCNN)}, pages 1--8, 2022.

\bibitem{qscan}
Rapha{\"e}l Avalos.
\newblock Exploration and communication for partially observable collaborative
  multi-agent reinforcement learning.
\newblock In {\em Proceedings of the 21st International Conference on
  Autonomous Agents and Multiagent Systems}, pages 1829--1832, 2022.

\bibitem{COMA}
Jakob Foerster, Gregory Farquhar, Triantafyllos Afouras, Nantas Nardelli, and
  Shimon Whiteson.
\newblock Counterfactual multi-agent policy gradients.
\newblock In {\em Proceedings of the AAAI Conference on Artificial
  Intelligence}, volume~32, 2018.

\bibitem{MADDPG}
Ryan Lowe, Yi~I Wu, Aviv Tamar, Jean Harb, OpenAI Pieter~Abbeel, and Igor
  Mordatch.
\newblock Multi-agent actor-critic for mixed cooperative-competitive
  environments.
\newblock {\em Advances in neural information processing systems}, 30, 2017.

\bibitem{DGN}
Jiechuan Jiang, Chen Dun, Tiejun Huang, and Zongqing Lu.
\newblock Graph convolutional reinforcement learning.
\newblock {\em arXiv preprint arXiv:1810.09202}, 2018.

\bibitem{G2ANet}
Yong Liu, Weixun Wang, Yujing Hu, Jianye Hao, Xingguo Chen, and Yang Gao.
\newblock Multi-agent game abstraction via graph attention neural network.
\newblock In {\em Proceedings of the AAAI Conference on Artificial
  Intelligence}, volume~34, pages 7211--7218, 2020.

\bibitem{LA-QTransformer}
Tianze Zhou, Fubiao Zhang, Kun Shao, Kai Li, Wenhan Huang, Jun Luo, Weixun
  Wang, Yaodong Yang, Hangyu Mao, Bin Wang, et~al.
\newblock Cooperative multi-agent transfer learning with level-adaptive credit
  assignment.
\newblock {\em arXiv preprint arXiv:2106.00517}, 2021.

\bibitem{TarMAC}
Abhishek Das, Th{\'e}ophile Gervet, Joshua Romoff, Dhruv Batra, Devi Parikh,
  Mike Rabbat, and Joelle Pineau.
\newblock Tarmac: Targeted multi-agent communication.
\newblock In {\em International Conference on Machine Learning}, pages
  1538--1546. PMLR, 2019.

\bibitem{CommNet}
Sainbayar Sukhbaatar, Rob Fergus, et~al.
\newblock Learning multiagent communication with backpropagation.
\newblock {\em Advances in neural information processing systems}, 29, 2016.

\bibitem{AI-QMIX}
Shariq Iqbal, Christian A.~Schröder de~Witt, Bei Peng, Wendelin Böhmer,
  Shimon Whiteson, and Fei Sha.
\newblock Ai-qmix: Attention and imagination for dynamic multi-agent
  reinforcement learning.
\newblock {\em CoRR}, abs/2006.04222, 2020.

\bibitem{DCG}
Wendelin Boehmer, Vitaly Kurin, and Shimon Whiteson.
\newblock Deep coordination graphs.
\newblock In Hal~Daumé III and Aarti Singh, editors, {\em Proceedings of the
  37th International Conference on Machine Learning}, volume 119 of {\em
  Proceedings of Machine Learning Research}, pages 980--991. PMLR, 13--18 Jul
  2020.

\bibitem{GraphNets}
Peter~W. Battaglia, Jessica~B. Hamrick, Victor Bapst, Alvaro Sanchez-Gonzalez,
  Vinicius Zambaldi, Mateusz Malinowski, Andrea Tacchetti, David Raposo, Adam
  Santoro, Ryan Faulkner, Caglar Gulcehre, Francis Song, Andrew Ballard, Justin
  Gilmer, George Dahl, Ashish Vaswani, Kelsey Allen, Charles Nash, Victoria
  Langston, Chris Dyer, Nicolas Heess, Daan Wierstra, Pushmeet Kohli, Matt
  Botvinick, Oriol Vinyals, Yujia Li, and Razvan Pascanu.
\newblock Relational inductive biases, deep learning, and graph networks, 2018.

\bibitem{EdgeConv}
Yue Wang, Yongbin Sun, Ziwei Liu, Sanjay~E Sarma, Michael~M Bronstein, and
  Justin~M Solomon.
\newblock Dynamic graph cnn for learning on point clouds.
\newblock {\em Acm Transactions On Graphics (tog)}, 38(5):1--12, 2019.

\bibitem{DeepSets}
Manzil Zaheer, Satwik Kottur, Siamak Ravanbakhsh, Barnabas Poczos, Russ~R
  Salakhutdinov, and Alexander~J Smola.
\newblock Deep sets.
\newblock In I.~Guyon, U.~V. Luxburg, S.~Bengio, H.~Wallach, R.~Fergus,
  S.~Vishwanathan, and R.~Garnett, editors, {\em Advances in Neural Information
  Processing Systems}, volume~30. Curran Associates, Inc., 2017.

\bibitem{Kolmogorov}
Andrei~Nikolaevich Kolmogorov.
\newblock On the representation of continuous functions of many variables by
  superposition of continuous functions of one variable and addition.
\newblock In {\em Doklady Akademii Nauk}, volume 114, pages 953--956. Russian
  Academy of Sciences, 1957.

\bibitem{pnorm}
Caglar Gulcehre, Kyunghyun Cho, Razvan Pascanu, and Yoshua Bengio.
\newblock Learned-norm pooling for deep feedforward and recurrent neural
  networks.
\newblock In {\em Joint European Conference on Machine Learning and Knowledge
  Discovery in Databases}, pages 530--546. Springer, 2014.

\bibitem{DQN}
Volodymyr Mnih, Koray Kavukcuoglu, David Silver, Andrei~A Rusu, Joel Veness,
  Marc~G Bellemare, Alex Graves, Martin Riedmiller, Andreas~K Fidjeland, Georg
  Ostrovski, et~al.
\newblock Human-level control through deep reinforcement learning.
\newblock {\em nature}, 518(7540):529--533, 2015.

\bibitem{Pymarl2}
Jian Hu, Siyang Jiang, Seth~Austin Harding, Haibin Wu, and Shih wei Liao.
\newblock Rethinking the implementation tricks and monotonicity constraint in
  cooperative multi-agent reinforcement learning.
\newblock 2021.

\bibitem{SMAC}
Mikayel Samvelyan, Tabish Rashid, Christian~Schroeder de~Witt, Gregory
  Farquhar, Nantas Nardelli, Tim G.~J. Rudner, Chia-Man Hung, Philiph H.~S.
  Torr, Jakob Foerster, and Shimon Whiteson.
\newblock {The} {StarCraft} {Multi}-{Agent} {Challenge}.
\newblock {\em CoRR}, abs/1902.04043, 2019.

\bibitem{CSP}
Tomasz Michalak, Talal Rahwan, Edith Elkind, Michael Wooldridge, and
  Nicholas~R. Jennings.
\newblock A hybrid exact algorithm for complete set partitioning.
\newblock {\em Artificial Intelligence}, 230:14--50, 2016.

\bibitem{CSP-NPhard}
Gerhard~J. Woeginger.
\newblock {\em Exact Algorithms for NP-Hard Problems: A Survey}, pages
  185--207.
\newblock Springer Berlin Heidelberg, Berlin, Heidelberg, 2003.

\bibitem{Shapley}
Lloyd~S Shapley.
\newblock {\em 17. A value for n-person games}.
\newblock Princeton University Press, 2016.

\bibitem{KolmogorovSimplification}
G.~G. Lorentz.
\newblock Metric entropy, widths, and superpositions of functions.
\newblock {\em The American Mathematical Monthly}, 69(6):469--485, 1962.

\end{thebibliography}

\end{document}